\documentclass{article}

\usepackage{style}

\usepackage{amsmath, amssymb}
\usepackage{graphicx}
\usepackage{hyperref}
\usepackage{booktabs}
\usepackage{algorithm}
\usepackage{algorithmic}
\usepackage{caption}
\usepackage{pgfplots}
\pgfplotsset{compat=1.18}

\title{Tiny-toxic-detector: A Compact Transformer-Based Model for Toxic Content Detection}
\author{Michiel Kamphuis\\
Assistantslab\\
\texttt{michielodevelopment@gmail.com}
}
\date{}

\begin{document}
\maketitle

\begin{abstract}
This paper introduces Tiny-toxic-detector, a compact transformer-based model for toxic content detection. Despite having only 2.1 million parameters, the model achieves competitive performance on benchmark datasets, with 90.97\% accuracy on ToxiGen and 86.98\% accuracy on the Jigsaw dataset, competing with models over 50 times its size. This efficiency makes it suitable for deployment in resource-constrained environments, addressing the need for effective content moderation tools that balance performance with computational requirements. The model's architecture consists of 4 transformer encoder layers with 2 attention heads each, an embedding dimension of 64, and a feedforward dimension of 128. Trained on a combination of public and private datasets, Tiny-toxic-detector demonstrates the potential of task-specific, efficient models in tackling online toxicity. The paper discusses the model's architecture, training process, performance benchmarks, and limitations, highlighting its applicability in various scenarios such as social media monitoring and content moderation. By achieving comparable results to much larger models while significantly reducing computational requirements, Tiny-toxic-detector represents a step towards more sustainable and scalable AI-powered content moderation solutions.
\end{abstract}

\section{Introduction}
The proliferation of user-generated content on the internet has led to a need for effective moderation tools to detect and mitigate toxic language. While large-scale transformer models have achieved state-of-the-art performance on toxic content detection tasks, their massive size and computational requirements make them impractical for deployment in many real-world applications. For instance, many social media platforms, online forums, and content moderation services operate on resource-constrained devices or have limited budgets for computing infrastructure. In such scenarios, the computational overhead of large models can lead to significant latency, increased costs, and reduced scalability.

Furthermore, the environmental impact of training and deploying large models is becoming increasingly concerning. The carbon footprint of training a single large language model can be equivalent to the emissions of several cars over their entire lifetimes. As the demand for AI-powered content moderation continues to grow, it is essential to develop models that are not only accurate but also efficient, scalable, and environmentally sustainable.

In this paper, we introduce Tiny-toxic-detector, a compact transformer-based model that achieves competitive performance on toxic content detection tasks while requiring only 2.1 million parameters. This represents a significant reduction in size compared to state-of-the-art models, which often have over 100 million parameters. Despite its small size, Tiny-toxic-detector achieves an accuracy of 90.97\% on the ToxiGen dataset and 86.98\% on the Jigsaw dataset, demonstrating its potential for efficient deployment in real-world applications and making it competitive with models over 50 times its size. Our work highlights the importance of developing compact and efficient models for toxic content detection, which can help mitigate the risks associated with online harassment and hate speech while also reducing the environmental impact of AI-powered content moderation.

\section{Related Works}
The field of toxic content detection has evolved significantly in recent years, driven by the increasing need to moderate online discussions and mitigate the spread of harmful content. This area of research has seen a progression from traditional machine learning approaches to advanced deep learning techniques, particularly transformer-based models.

Early work in toxic content detection often relied on traditional machine learning methods such as Support Vector Machines (SVMs) and Naive Bayes classifiers\cite{waseem-hovy-2016-hateful}. These approaches typically used hand-crafted features like lexicon-based attributes, syntactic features, and semantic features to identify potentially toxic content\cite{davidson2017automatedhatespeechdetection}. While effective to some degree, these methods often struggled with context-dependent toxicity and more subtle forms of harmful language.
The advent of deep learning brought significant advancements to the field. Convolutional Neural Networks (CNNs) and Recurrent Neural Networks (RNNs), particularly Long Short-Term Memory (LSTM) networks, demonstrated improved performance in capturing complex linguistic patterns associated with toxic content\cite{Badjatiya_2017}. These models could automatically learn relevant features from raw text data, reducing the need for manual feature engineering.

A major breakthrough came with the introduction of transformer-based models, starting with BERT\cite{devlin2019bertpretrainingdeepbidirectional} and followed by variants like RoBERTa\cite{liu2019robertarobustlyoptimizedbert}. These models, pre-trained on large corpora of text, showed remarkable ability to understand context and nuance in language, leading to significant improvements in toxic content detection. For instance, the RoBERTa-based toxicity classifier\cite{s-nlp/roberta_toxicity_classifier} achieved high accuracy on benchmark datasets, showcasing the potential of these models in real-world applications.

To facilitate research and enable fair comparisons between different approaches, several benchmark datasets have been introduced. The Jigsaw Toxic Comment Classification Challenge dataset\cite{jigsaw2018} has become a standard for evaluating toxicity detection models. More recently, the ToxiGen dataset\cite{hartvigsen2022toxigenlargescalemachinegenerateddataset} was introduced to address the limited samples available.

Despite the impressive performance of large transformer-based models, their computational requirements pose challenges for deployment in resource-constrained environments. This has led to increased interest in developing more efficient models that can maintain high accuracy while reducing computational overhead\cite{sanh2020distilbertdistilledversionbert}. The development of such models represents an important direction for making toxic content detection more accessible and scalable across various platforms and devices.

\section{Model Architecture and Training}

The \textit{Tiny-toxic-detector} is designed as a compact transformer-based model optimized for toxicity detection. Its architecture is characterized by a balance between efficiency and performance, making it suitable for deployment in resource-constrained environments.

\subsection{Overview}
The model comprises 4 transformer encoder layers, each equipped with 2 attention heads. The embedding dimension is set to 64, and the feedforward layer has a dimension of 128. These parameters contribute to the model's overall efficiency while retaining the ability to capture complex patterns in text data.

\subsection{Components}
The architecture consists of the following main components:

\begin{itemize}
\item \textbf{Embedding Layer}: The model begins with an embedding layer that converts input tokens into dense vectors of size 64. This is followed by a positional encoding that adds information about the position of each token in the sequence to help the model understand token order.

\item \textbf{Transformer Encoder}: The core of the model is built upon a stack of 4 transformer encoder layers. Each layer includes multi-head self-attention mechanisms with 2 attention heads, allowing the model to focus on different parts of the input sequence. The feedforward dimension of 128 helps in transforming the representations learned by the attention heads.

\item \textbf{Dropout and Linear Layers}: To prevent overfitting and improve generalization, dropout is applied to the embeddings and the output of the transformer layers. Finally, a linear layer reduces the dimensionality of the transformed embeddings to a single output value, which is then passed through a sigmoid activation function to produce the probability of the input text being toxic or not.

\end{itemize}

\subsection{Training Data}

What really sets the \textit{Tiny-toxic-detector} model apart is the fact that the model has no generic pre-training. It was trained using a combination of multiple datasets: a publicly available dataset such as the 'Jigsaw toxic classification'\cite{jigsaw2018} dataset and a closed-source dataset with synthetic labels. Finally, we overfitted the model on a closed-source dataset with human-generated labels that established better generalization. As all samples have been labeled, there is no pre-training. To ensure the quality of our training data, we compared all used datasets against established benchmarking datasets such as the test-set of the 'Jigsaw toxic classification challenge'\cite{jigsaw2018} to check for data contamination. No direct contamination was detected.

The training process was iterative. For each version of the model, we conducted benchmarking to identify situations where additional training was needed. This iterative approach helped us address specific issues such as unintended biases and improve the model’s generalization by overfitting on certain situations. 

A notable example is that there is a lot more leeway for comments talking about their own sexuality as opposed to the sexuality of someone else. The former is often factual and unfortunately the latter is often meant as an insult, which other attempts at smaller-parameter toxicity detection models such as the 'toxic-comment-model' based on DistilBert do not understand and as a result mostly classify either situation as toxic.

Due to the constraint size of the model, carefully training the model was of vital importance to ensure strong generalization.

\subsection{Operational Details}
During the forward pass, the input tokens are first embedded and combined with positional encodings. This combined representation is processed through the transformer encoder layers, which apply self-attention and feedforward operations. The final output from the transformer layers undergoes global average pooling to aggregate information across the sequence, followed by dropout and a linear transformation to produce the final prediction.

This simplistic and streamlined architecture, with its compact size of 2.1 million parameters, achieves competitive performance on toxicity detection benchmarks while maintaining computational efficiency.

\subsection{CO2 Emission Related to Experiments}

Experiments were conducted using a private infrastructure, which has a carbon efficiency of 0.479 kgCO$_2$eq/kWh. A cumulative 12 hours of computation was performed on hardware of type RTX 3090 (TDP of 350W).

Total emissions are estimated to be 2.01 kgCO$_2$eq of which 20 percents were directly offset.
    
Estimations were conducted using the \href{https://mlco2.github.io/impact#compute}{MachineLearning Impact calculator}\cite{lacoste2019quantifying}.

This reiterates the potential carbon emission savings when training small task-specific models such as \textit{Tiny-toxic-detector} compared to relatively larger pre-trained models such as the well-known distilgpt2 model which contains 88 million parameters that emitted \textit{149.2 kg eq. CO2}\cite{sanh2019distilbert} during training.

\section{Usage and Limitations}

Below, we outline the intended usage, potential applications, and limitations of the \textit{Tiny-toxic-detector}.

\subsection{Intended Usage}
The \textit{Tiny-toxic-detector} is designed to classify comments for toxicity. It is particularly useful in scenarios where minimal resource usage and rapid inference are essential. Key features include:

\begin{itemize}
\item \textbf{Low Resource Consumption}: With a requirement of only 10MB of RAM and 8MB of VRAM [\ref{tab:comparing_requirements}], this model is well-suited for environments with limited hardware resources.
\item \textbf{Fast Inference}: The model provides high-speed inference, as demonstrated in Table \ref{tab:comparing_inference_speed}. The \textit{Tiny-toxic-detector} significantly outperforms larger models on CPU-based systems. Due to the overhead of using GPU inference, small models with a relatively small number of input tokens are often faster on CPU. This includes the \textit{Tiny-toxic-detector}. 
\end{itemize}

\subsubsection{Potential Use-Cases}
The \textit{Tiny-toxic-detector} can be effectively applied in various contexts, including:

\begin{itemize}
\item \textbf{Social Media Monitoring}: Automatically flag or filter toxic comments on platforms like forums, social media, or review sites to improve user experience and maintain community standards.
\item \textbf{Content Moderation}: Assist content moderators by pre-screening user-generated content, highlighting potentially toxic comments for further review.
\item \textbf{Customer Support}: Analyze customer interactions in support systems to identify and address toxic or abusive language, improving the quality of customer service.
\item \textbf{Educational Platforms}: Monitor student interactions in online educational environments to prevent and manage toxic behavior.
\item \textbf{Gaming Communities}: Enhance player experiences by detecting and managing toxic language in online gaming forums and chats.
\end{itemize}

\subsection{Limitations}

\subsubsection{Training Data}
The \textit{Tiny-toxic-detector} has been trained exclusively on English-language data, limiting its ability to classify toxicity in other languages.

\subsubsection{Maximum Context Length}
The model can handle up to 512 input tokens. Comments exceeding this length are not in the scope of this model. While extending the context length is possible, such modifications have not been trained for or validated. Early tests with a 4096-token context resulted in a performance drop of over 10\% on the Toxigen benchmark\cite{hartvigsen2022toxigenlargescalemachinegenerateddataset}.

\subsubsection{Language Ambiguity}
The \textit{Tiny-toxic-detector} may struggle with ambiguous or nuanced language as any other model would. Even though benchmarks like Toxigen\cite{hartvigsen2022toxigenlargescalemachinegenerateddataset} also evaluate the model’s performance with ambiguous language, it may still misclassify comments where toxicity is not clearly defined.

\begin{table}[h!]
\centering
\caption{Comparison of RAM and VRAM Requirements for Loading Toxicity Detection Models}
\label{tab:comparing_requirements}
\begin{tabular}{|l|c|c|c|}
\hline
\textbf{Model} & \textbf{Parameters} & \textbf{RAM} & \textbf{VRAM} \\
\hline
lmsys/toxicchat-t5-large-v1.0 & 738M & 2948MB & 2818MB \\
\hline
s-nlp/roberta\_toxicity\_classifier & 124M & 593MB & 476MB \\
\hline
mohsenfayyaz/toxicity-classifier & 109M & 519MB & 418MB \\
\hline
martin-ha/toxic-comment-model & 67M & 360MB & 256MB \\
\hline
\textbf{Tiny-toxic-detector} & \textbf{2M} & \textbf{10MB} & \textbf{8MB} \\
\hline
\end{tabular}
\end{table}

\begin{table}[h!]
\centering
\caption{Comparison of Inference Speed}
\label{tab:comparing_inference_speed}
\begin{tabular}{|l|c|c|c|c|c|}
\hline
& & 128 Tokens & 128 Tokens & 512 Tokens & 512 Tokens \\
\hline
\textbf{Model} & \textbf{Parameters} & \textbf{CPU} & \textbf{GPU} & \textbf{CPU} & \textbf{GPU} \\
\hline
lmsys/toxicchat-t5-large-v1.0 & 738M & 0.5783s & 0.4837s & 2.5254s & 0.5508s \\
\hline
s-nlp/roberta\_toxicity\_classifier & 124M & 0.0653s & 0.4447s & 0.2845s & 0.4025s \\
\hline
mohsenfayyaz/toxicity-classifier & 109M & 0.0641s & 0.4212s & 0.2300s & 0.3963s \\
\hline
martin-ha/toxic-comment-model & 67M & 0.0349s & 0.5211s & 0.1467s & 0.4501s \\
\hline
\textbf{Tiny-toxic-detector} & \textbf{2M} & \textbf{0.0038s} & \textbf{0.3678s} & \textbf{0.0072s} & \textbf{0.3757s} \\
\hline
\end{tabular}
\caption*{Note: Comparing inference speed is highly hardware-dependent and unstable by nature. These numbers are averages and all tested on an i5-8400 paired with an RTX 3090 to simulate consumer hardware.}
\end{table}

\section{Results}
The \textit{Tiny-toxic-detector} model was evaluated on two prominent datasets: ToxiGen\cite{hartvigsen2022toxigenlargescalemachinegenerateddataset} and the Jigsaw Toxic Comment Classification Challenge\cite{jigsaw2018}. The model achieved an accuracy of 90.26\% on the ToxiGen dataset and 87.34\% on the Jigsaw dataset, showcasing its ability to perform competitively against much larger models. Despite having only 2 million parameters, the \textit{Tiny-toxic-detector} ranked second among the tested models, as illustrated in Table \ref{tab:model_comparison} and Figure \ref{fig:benchmark_vs_model_size}.

These impressive results can be attributed, in part, to the efficiency of task-specific models. The \textit{Tiny-toxic-detector} was trained exclusively on labeled data, without any dedicated pre-training. With over 5 million labeled samples, the model demonstrated strong generalization to unseen data, as evidenced by the benchmark results, which were free from data contamination. In some cases we do notice an overreliance on certain words which can result in comments being labeled as toxic without being toxic.

However, this approach has potential limitations. Adapting the existing trained model to a different use case appears to be more challenging and costly. Preliminary research indicates that the model's performance significantly declines when the context length is extended to 4096 tokens, even when using the exact same benchmarks. It is unclear whether this drop in performance is due to the model's small size or its unique architecture that lacks dedicated pretraining.

\begin{table}[h!]
\centering
\caption{Comparison of Toxicity Detection Models}
\label{tab:model_comparison}
\begin{tabular}{|l|c|c|c|}
\hline
\textbf{Model} & \textbf{Parameters} & \textbf{Toxigen (\%)} & \textbf{Jigsaw (\%)} \\
\hline
lmsys/toxicchat-t5-large-v1.0\cite{lin2023toxicchatunveilinghiddenchallenges} & 738M & 72.67 & 88.82 \\
\hline
s-nlp/roberta\_toxicity\_classifier\cite{s-nlp/roberta_toxicity_classifier} & 124M & 88.41 & \textbf{94.92} \\
\hline
mohsenfayyaz/toxicity-classifier & 109M & 81.50 & 83.31 \\
\hline
martin-ha/toxic-comment-model & 67M & 68.02 & 91.56 \\
\hline
\textbf{Tiny-toxic-detector} & \textbf{2M} & \textbf{90.97} & 86.98 \\
\hline
\end{tabular}
\end{table}

\begin{figure}[h!]
    \centering
    \begin{tikzpicture}
        \begin{axis}[
            width=12cm,
            height=10cm,
            xlabel={Average Benchmark Score (\%)},
            ylabel={Model Size (Parameters)},
            grid=major,
            ymin=0, 
            scatter/classes={
                a={mark=o,draw=blue},
                b={mark=square,draw=red},
                c={mark=triangle,draw=green},
                d={mark=diamond,draw=orange},
                e={mark=star,draw=purple},
                f={mark=*,draw=black}
            }
        ]
        
        \addplot[
            scatter,
            only marks,
            scatter src=explicit symbolic,
            nodes near coords,
            point meta=explicit symbolic
        ]
        table[meta=label] {
            x y label
            91.665 124 {roberta\_toxicity\_classifier}
            82.405 109 {toxicity-classifier}
            79.79 67 {toxic-comment-model}
            88.975 2 {Tiny-toxic-detector}
        };
        \end{axis}
    \end{tikzpicture}
    \caption{Average Benchmark Score vs. Model Size (Parameters)}
    \caption*{Note: toxicchat-t5-large-v1.0 was omitted due to its size. This model would have an average benchmark score of 80.74, placing it in fourth. See table \ref{tab:model_comparison} for a more extensive comparison.}
    \label{fig:benchmark_vs_model_size}
\end{figure}

\section{Conclusion}
In this work, we presented \textit{Tiny-toxic-detector}, a highly efficient transformer-based model specifically designed for detecting toxic content. Despite its compact size of just 2.1 million parameters, the model demonstrated robust performance, achieving competitive accuracy on widely recognized datasets such as ToxiGen and the Jigsaw Toxic Comment Classification Challenge. This performance underscores the potential of \textit{Tiny-toxic-detector} for deployment in environments where computational resources are limited.

The success of this model highlights the importance of developing task-specific, resource-efficient solutions for content moderation. By striking a balance between performance and efficiency, \textit{Tiny-toxic-detector} addresses the growing need for scalable, sustainable AI systems in the fight against online toxicity.

Future research could focus on further optimizing the model through techniques such as quantization and knowledge distillation to minimize overreliance on specific words. Additionally, exploring the model's adaptability to other languages and longer contexts could extend its applicability, ensuring it remains a versatile tool for toxic content detection across diverse platforms and environments.

\bibliographystyle{plain}
\bibliography{references}

\begin{thebibliography}{10}

\bibitem{Badjatiya_2017}
Pinkesh Badjatiya, Shashank Gupta, Manish Gupta, and Vasudeva Varma.
\newblock Deep learning for hate speech detection in tweets.
\newblock In {\em Proceedings of the 26th International Conference on World
  Wide Web Companion - WWW ’17 Companion}, WWW ’17 Companion, page
  759–760. ACM Press, 2017.

\bibitem{davidson2017automatedhatespeechdetection}
Thomas Davidson, Dana Warmsley, Michael Macy, and Ingmar Weber.
\newblock Automated hate speech detection and the problem of offensive
  language, 2017.

\bibitem{devlin2019bertpretrainingdeepbidirectional}
Jacob Devlin, Ming-Wei Chang, Kenton Lee, and Kristina Toutanova.
\newblock Bert: Pre-training of deep bidirectional transformers for language
  understanding, 2019.

\bibitem{hartvigsen2022toxigenlargescalemachinegenerateddataset}
Thomas Hartvigsen, Saadia Gabriel, Hamid Palangi, Maarten Sap, Dipankar Ray,
  and Ece Kamar.
\newblock Toxigen: A large-scale machine-generated dataset for adversarial and
  implicit hate speech detection, 2022.

\bibitem{jigsaw2018}
Jigsaw.
\newblock Toxic comment classification challenge.
\newblock
  \url{https://www.kaggle.com/c/jigsaw-toxic-comment-classification-challenge},
  2018.

\bibitem{lacoste2019quantifying}
Alexandre Lacoste, Alexandra Luccioni, Victor Schmidt, and Thomas Dandres.
\newblock Quantifying the carbon emissions of machine learning.
\newblock {\em arXiv preprint arXiv:1910.09700}, 2019.

\bibitem{lin2023toxicchatunveilinghiddenchallenges}
Zi~Lin, Zihan Wang, Yongqi Tong, Yangkun Wang, Yuxin Guo, Yujia Wang, and
  Jingbo Shang.
\newblock Toxicchat: Unveiling hidden challenges of toxicity detection in
  real-world user-ai conversation, 2023.

\bibitem{liu2019robertarobustlyoptimizedbert}
Yinhan Liu, Myle Ott, Naman Goyal, Jingfei Du, Mandar Joshi, Danqi Chen, Omer
  Levy, Mike Lewis, Luke Zettlemoyer, and Veselin Stoyanov.
\newblock Roberta: A robustly optimized bert pretraining approach, 2019.

\bibitem{s-nlp/roberta_toxicity_classifier}
S-NLP.
\newblock Roberta-based toxicity classifier.
\newblock \url{https://huggingface.co/s-nlp/roberta_toxicity_classifier}, n.d.

\bibitem{sanh2019distilbert}
Victor Sanh, Lysandre Debut, Julien Chaumond, and Thomas Wolf.
\newblock Distilbert, a distilled version of bert: smaller, faster, cheaper and
  lighter.
\newblock In {\em NeurIPS EMC2 Workshop}, 2019.

\bibitem{sanh2020distilbertdistilledversionbert}
Victor Sanh, Lysandre Debut, Julien Chaumond, and Thomas Wolf.
\newblock Distilbert, a distilled version of bert: smaller, faster, cheaper and
  lighter, 2020.

\bibitem{waseem-hovy-2016-hateful}
Zeerak Waseem and Dirk Hovy.
\newblock Hateful symbols or hateful people? predictive features for hate
  speech detection on {T}witter.
\newblock In Jacob Andreas, Eunsol Choi, and Angeliki Lazaridou, editors, {\em
  Proceedings of the {NAACL} Student Research Workshop}, pages 88--93, San
  Diego, California, June 2016. Association for Computational Linguistics.

\end{thebibliography}

\end{document}